\documentclass{article}

\usepackage[preprint]{neurips_2024}
\usepackage{amsmath} 
\usepackage{graphicx} 
\usepackage{tabularx} 
\usepackage{subfigure}

\usepackage[utf8]{inputenc} 
\usepackage[T1]{fontenc}    
\usepackage{hyperref}       
\usepackage{url}            
\usepackage{booktabs}       
\usepackage{amsfonts}       
\usepackage{nicefrac}       
\usepackage{microtype}      
\usepackage{xcolor}         

\newcommand{\systemname}{SpINR (Ours)}
\newcommand{\systemnamesansours}{SpINR}

\title{SpINR: Neural Volumetric Reconstruction for FMCW Radars}

\author{%
  Harshvardhan Takawale\\
  Department of Computer Science\\
  University of Maryland, College Park \\
  \texttt{htakawal@umd.edu} \\
  \And
  Nirupam Roy \\
  Department of Computer Science \\
  University of Maryland, College Park \\
  \texttt{niruroy@umd.edu} \\
}

\begin{document}

\maketitle

\begin{abstract}
In this paper, we introduce \textbf{\systemnamesansours{}}, a novel framework for volumetric reconstruction using Frequency-Modulated Continuous-Wave (FMCW) radar data. Traditional radar imaging techniques, such as backprojection, often assume ideal signal models and require dense aperture sampling, leading to limitations in resolution and generalization. To address these challenges, SpINR integrates a fully differentiable forward model that operates natively in the frequency domain with implicit neural representations (INRs). This integration leverages the linear relationship between beat frequency and scatterer distance inherent in FMCW radar systems, facilitating more efficient and accurate learning of scene geometry. Additionally, by computing outputs for only the relevant frequency bins, our forward model achieves greater computational efficiency compared to time-domain approaches that process the entire signal before transformation. Through extensive experiments, we demonstrate that SpINR significantly outperforms classical backprojection methods and existing learning-based approaches, achieving higher resolution and more accurate reconstructions of complex scenes. This work represents the first application of neural volumetric reconstruction in the radar domain, offering a promising direction for future research in radar-based imaging and perception systems.
\end{abstract}


\section{Introduction}

Radar-based sensing has emerged as a robust and scalable modality for 3D perception in scenarios where traditional optical or LiDAR systems struggle—such as in adverse weather, poor lighting, or through occlusions. Among radar technologies, Frequency-Modulated Continuous-Wave (FMCW) radar stands out for its ability to deliver high-resolution range and velocity measurements using compact, low-power hardware. As FMCW radars become increasingly prevalent in autonomous vehicles, robotics, and smart sensing, there is a growing need for accurate and efficient methods to reconstruct 3D structure from radar measurements.

However, traditional radar imaging techniques such as backprojection, range-Doppler processing, and voxelized synthetic aperture radar (SAR) reconstructions suffer from two core limitations. First, these methods rely on oversimplified or idealized physical models, often neglecting the spectral characteristics of radar signals such as frequency bin mismatch, side-lobes, and spillage due to discretization. Second, the spatial resolution of reconstructions is bottlenecked by voxel grid resolution and uniform sampling assumptions, making high-fidelity reconstruction under sparse or irregular apertures particularly challenging.

To address these issues, we propose \textbf{\systemnamesansours}, a novel radar imaging framework that unifies physically grounded signal modeling with the expressivity of implicit neural representations (INRs). Our key insight is to operate directly in the \textit{frequency domain}, leveraging the linear relationship between beat frequency and scatterer distance in FMCW radar. This allows us to formulate a closed-form, differentiable forward model that analytically computes the radar response at a sparse set of frequency bins—bypassing the need for time-domain simulations and expensive FFT post-processing. We then supervise an implicit neural field to reconstruct volumetric geometry that matches these observed frequency-domain radar signals across multiple transmit-receive pairs and poses. Our approach offers multiple advantages over classical and learning-based baselines: \\
    (1) \textbf{Frequency-Domain Forward Modeling:} By analytically computing the radar frequency response, our method avoids numerical approximations such as time-domain convolution and FFT truncation, leading to faster and more accurate supervision. \\
    (2) \textbf{Continuous Volumetric Representation:} Using neural fields instead of voxel grids allows our method to generalize to complex geometries and sparsely sampled synthetic apertures while maintaining sub-voxel precision. \\
    (3) \textbf{Differentiable and Compact Formulation:} Our forward model is closed-form and fully differentiable, making it suitable for gradient-based optimization and scalable integration into modern rendering pipelines. 

We conduct extensive experiments across multiple 3D objects and compare against classical backprojection, time-domain forward models, and time-domain supervision baselines. Our method consistently achieves higher fidelity—demonstrated via SSIM, Chamfer Distance, PSNR, LPIPS, and IoU—while also reducing the computational overhead of forward modeling.

This work presents the first end-to-end framework for neural volumetric reconstruction from FMCW radar signals using a closed-form frequency-domain forward model. We introduce a differentiable radar signal model that accurately accounts for spectral effects and enables direct learning in the frequency domain. Our approach achieves state-of-the-art performance in 3D reconstruction with strong implications for neural modeling of real-world RF sensing. 


\begin{figure}[t]
  \centering


  \includegraphics[width=\linewidth]{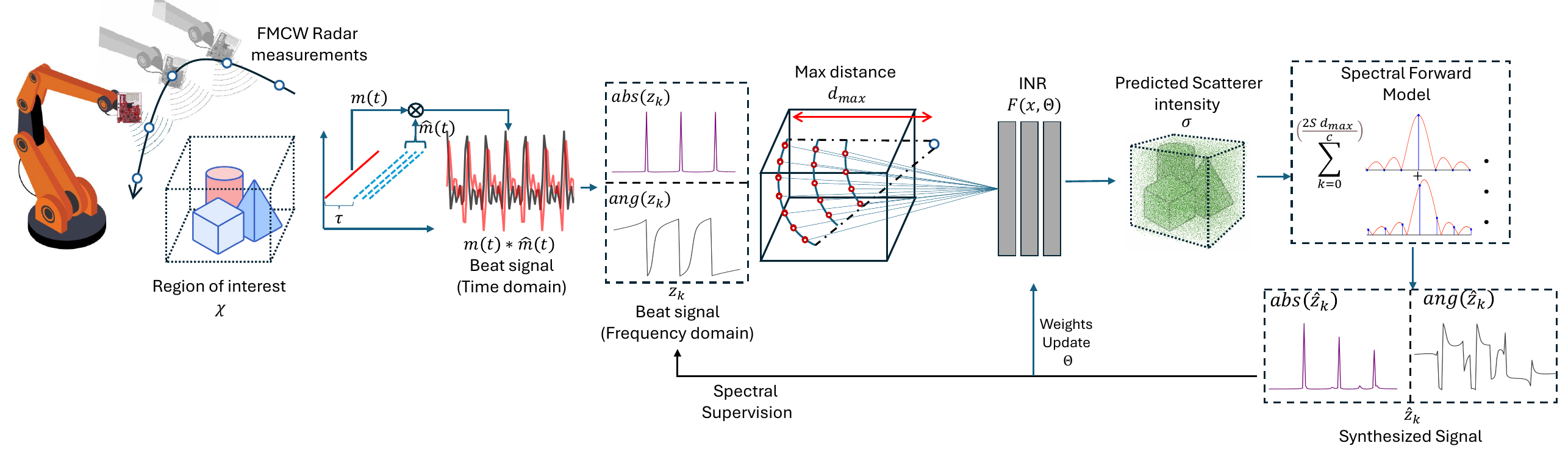}

  \caption{Overview of our method. FMCW radar measurements are collected from multiple viewpoints and transformed to the frequency domain for supervision. An implicit neural representation (INR) predicts scatterer intensity over the scene, and our differentiable spectral forward model synthesizes the complex-valued frequency-domain signal (magnitude and phase). The INR is trained by minimizing the discrepancy between synthesized and measured frequency-domain signals, using only the bins corresponding to the valid scene distances.}

  \label{fig:overview}
\end{figure}


\section{Related Work}

\subsection{FMCW Radar Imaging and Signal Modeling}

FMCW (Frequency-Modulated Continuous-Wave) radar systems are widely used for active sensing in automotive, industrial, and robotics applications due to their ability to simultaneously estimate range and velocity from compact, low-power hardware~\cite{7870764,6247496,10.1002/rob.20393}. The key signal processing pipeline involves mixing the reflected chirp with the transmitted signal to produce a beat frequency, which is then analyzed via FFT to infer round-trip delay. This enables high-resolution range imaging via spectral estimation.

Prior works in radar imaging typically employ classical reconstruction techniques such as range-Doppler processing~\cite{wagner2013wide}, backprojection~\cite{767270}, or synthetic aperture radar (SAR)~\cite{6504845}, often assuming ideal signal models and requiring dense aperture sampling. More recent works have explored inverse methods using deep learning, including radar-to-depth regression~\cite{xu2022learned}, radar occupancy estimation~\cite{sonny2024dynamic}, and neural SAR imaging~\cite{HUANG2020179}. However, these approaches often operate in the time domain or rely on voxelized grid representations, limiting their resolution and generalization.

Our work differs in two key ways: (1) we introduce a fully differentiable forward model for FMCW radar that operates natively in the frequency domain, and (2) we integrate this model into an implicit neural representation to enable high-fidelity volumetric reconstruction. This allows us to account for spectral effects such as frequency-bin mismatch and leakage, which are often ignored in time-domain approximations.

\subsection{Learning-Based Radar Imaging}

Several recent works have explored data-driven radar reconstruction pipelines using deep learning. These include radar-to-depth regression~\cite{9506550}, occupancy estimation~\cite{2405.13307}, and learned beamforming~\cite{al2022review}. While these methods have shown success in specific tasks, most rely on time-domain signal modeling and voxelized representations, limiting their resolution and physical fidelity.

In contrast, we introduce a frequency-domain forward model that captures spectral structure analytically and integrates seamlessly with continuous implicit neural representations.

\subsection{Implicit Neural Representations}
Implicit neural representations (INRs) or neural fields model continuous signals (e.g., 3D scenes, audio, or volumetric data) using neural networks that map coordinates to signal values. Pioneered by methods like NeRF~\cite{10.1145/3503250}, INRs have been widely adopted for view synthesis, shape reconstruction, and signal modeling~\cite{10.5555/3495724.3496350}. Recent extensions have applied INRs beyond RGB imagery to non-visual domains such as MRI~\cite{2407.02744}, acoustic~\cite{10.1145/3592141},  ultrasound~\cite{chen2024neural}, and transient photon imaging~\cite{malik2023transient}. These works leverage differentiable forward models and physically grounded supervision to reconstruct spatial or temporal fields from sparse measurements.

Our work brings INRs into the FMCW radar domain, where we learn a continuous volumetric representation optimized to match observed radar frequency-domain responses. To the best of our knowledge, this is the first work to combine INRs with a radar-specific, frequency-domain differentiable forward model.

\subsection{MIMO Radar and Synthetic Apertures}
MIMO radar systems enhance angular resolution by using virtual arrays created from multiple transmit-receive (Tx-Rx) pairs. Combined with synthetic aperture sampling, they enable dense 3D imaging from a compact sensor setup. Prior works often process MIMO data via backprojection or time-domain matched filtering~\cite{9136646}, which requires dense uniform sampling and lacks differentiability. While some studies simulate radar data for training~\cite{malik2023transient}, they do not integrate a compact forward model that captures frequency-domain behavior.

Our setup emulates a cylindrical inverse synthetic aperture via actuator and turntable motion and supports non-uniform sampling. We simulate a MIMO radar similar to AWR1843BOOST and apply a multistatic-to-monostatic conversion. The result is a sparse, yet physically accurate dataset used to supervise our differentiable frequency-domain model.

\subsection{Limitations of Backprojection Methods}

Backprojection is widely used in radar and CT imaging due to its physical interpretability~\cite{duersch2013backprojection}, but it has critical limitations: (1) it relies on uniform spatial discretization, leading to aliasing under sparse sampling; and (2) it is non-differentiable, making it unsuitable for learning-based pipelines.

Our frequency-domain forward model addresses these limitations by explicitly modeling spectral structure and enabling gradient-based optimization of a continuous volumetric field. This leads to more accurate, high-resolution reconstructions, particularly under sparse and irregular sampling.

\section{Formulation of Frequency-Domain Forward Model}
\subsection{Primer: FMCW Chirp Signal}
We develop our system model based on the mmWave radar sensors that transmits a signal with linearly time varying frequency. Such a signal is known as the Frequency modulated continuous wave signal (FMCW) and for a transceiver at location $(x_t,y_t,z_t)$ the signal can be formulated as - 
\[
m(t) = e^{\left(j2\pi\left(f_0 t + 0.5 S t^2\right)\right)}, \quad 0 \leq t \leq T,
\]
where $f0$ is the start frequency of the chirp and $S$ is the slope defined with units $Hz/s$. If there is a scatterer present at $(x_0,y_0,z_0)$ and reflects the signal with intensity $\sigma$ the received signal is delayed due to the distance traveled and scaled due to the propagation loss and $\sigma$ -
\[
\hat m(t) = \sigma m(t-\tau)  = \sigma e^{\left(j2\pi\left(f_0 (t-\tau) + 0.5 S (t-\tau)^2\right)\right)}, \quad 0 \leq t \leq T,
\]

where $\tau$ is the round-trip delay. If the distance between $(x_t,y_t,z_t)$ and $(x_0,y_0,z_0)$ is $\sqrt{(x_t - x_0)^2 + (y_t - y_0)^2 + (z_t - z_0)^2} = d$ $\tau$ is defined as $2d/C$. 

The FMCW radar mixes the transmitted and the received signal to demodulate the signal to a lower band. This is known as dechirping. The created beat signal is defined as - 
\begin{equation*}
m(t)* \hat m(t) = e^{(j2\pi f_0\tau + St\tau - 0.5Kt^2)}
\end{equation*}

the $-0.5Kt^2$ term is known as the residual video phase (RVP) and is often ignored as it is negligible~\cite{6810197} which simplifies the beat signal equation to-

\begin{align*}
m(t) * \hat{m}(t) = b(t) &= e^{j 2\pi (f_0 + S t)\tau} \nonumber \\
                  &= e^{j 2\pi f_0 \tau} \cdot e^{j 2\pi (S \tau) t}
\end{align*}

\subsubsection{Range estimation and resolution for FMCW Radar}

As derived above, the frequency of the beat signal is $f_b = S\tau = S(2d/c)$. This indicates a linear relationship between the range of a scatterer and the frequency of the beat signal. Therefore, applying an FFT to the beat signal results in a frequency-domain representation where each frequency bin corresponds to a specific range — a process referred to as range compression. This transforms time-delay information into a frequency domain representation, enabling efficient range estimation.

The ability of the radar to resolve two targets at different ranges — i.e., range resolution — is determined by the frequency resolution of the FFT, which is $\Delta f = \frac{1}{T_c}$.
Using the relationship $f_b = 2Sd/c$, the frequency resolution translates into a range resolution:
\[
\Delta f = \frac{2S\Delta d}{c} \quad \Rightarrow \quad \Delta d = \frac{c}{2S} \cdot \Delta f = \frac{c}{2B}
\]
Hence, the range resolution $\Delta d$ depends only on the bandwidth $B$ of the chirp and is independent of the chirp duration $T_c$ or the starting frequency $f_0$. To simplify analysis, we set $f_0 = 0$ without loss of generality, as it does not affect the range resolution. Under this simplification, the beat signal becomes:
\[
b(t) = e^{j2\pi (S t) \tau}
\]
This simplified expression retains the essential range information and is used in our training pipeline for range-aware signal representation.

\subsection{Spectral Transformation}
\label{DFT}

In practical radar systems, signals are sampled and finite in duration. As such, spectral analysis is performed using the Discrete Fourier Transform (DFT), which approximates the Discrete-Time Fourier Transform (DTFT) at uniformly spaced frequency bins $\beta_k = \frac{2\pi k}{N}$.

To understand how DFT distributes energy across bins, consider a complex exponential signal $S_t = M e^{i(\alpha t + \phi)}$ with arbitrary angular frequency $\alpha$. The DFT evaluates to:
\[
Z_k = \frac{M}{N} e^{i\phi} \sum_{t=0}^{N-1} e^{i(\alpha - \beta_k)t}
= \frac{M}{N} e^{i\phi} \cdot \frac{1 - e^{i(\alpha - \beta_k)N}}{1 - e^{i(\alpha - \beta_k)}}
\]

This formulation can be rewritten in closed form by noting that $\beta_k N = 2\pi k$:
\[
\boxed{
Z_k = \frac{M}{N} e^{i\phi} \cdot \frac{1 - e^{i \alpha N}}{1 - e^{i(\alpha - \beta_k)}}
}
\]

This expression illustrates how energy at frequency $\alpha$ leaks into neighboring bins unless it perfectly aligns with a DFT bin center $\beta_k$. The extent of this leakage is explored next.

\subsubsection{Spectral Leakage and Energy Spillage}

When $\alpha = \beta_{\hat{k}}$, all energy falls into bin $\hat{k}$: $Z_k = M e^{i\phi}$ if $k = \hat{k}$, and zero otherwise—which means there is no leakage to other bins.

In all other cases, the energy is not localized and spreads across bins. The resulting magnitude spectrum follows the Dirichlet kernel:
\[
|Z_k| \propto \left| \frac{\sin\left(\frac{N}{2}(\alpha - \beta_k)\right)}{\sin\left(\frac{1}{2}(\alpha - \beta_k)\right)} \right|
\]
which resembles a sinc-like envelope centered at $\alpha$.

\begin{figure}[h]
  \centering
  \begin{minipage}[b]{0.45\linewidth}
    \centering
    \includegraphics[width=\linewidth]{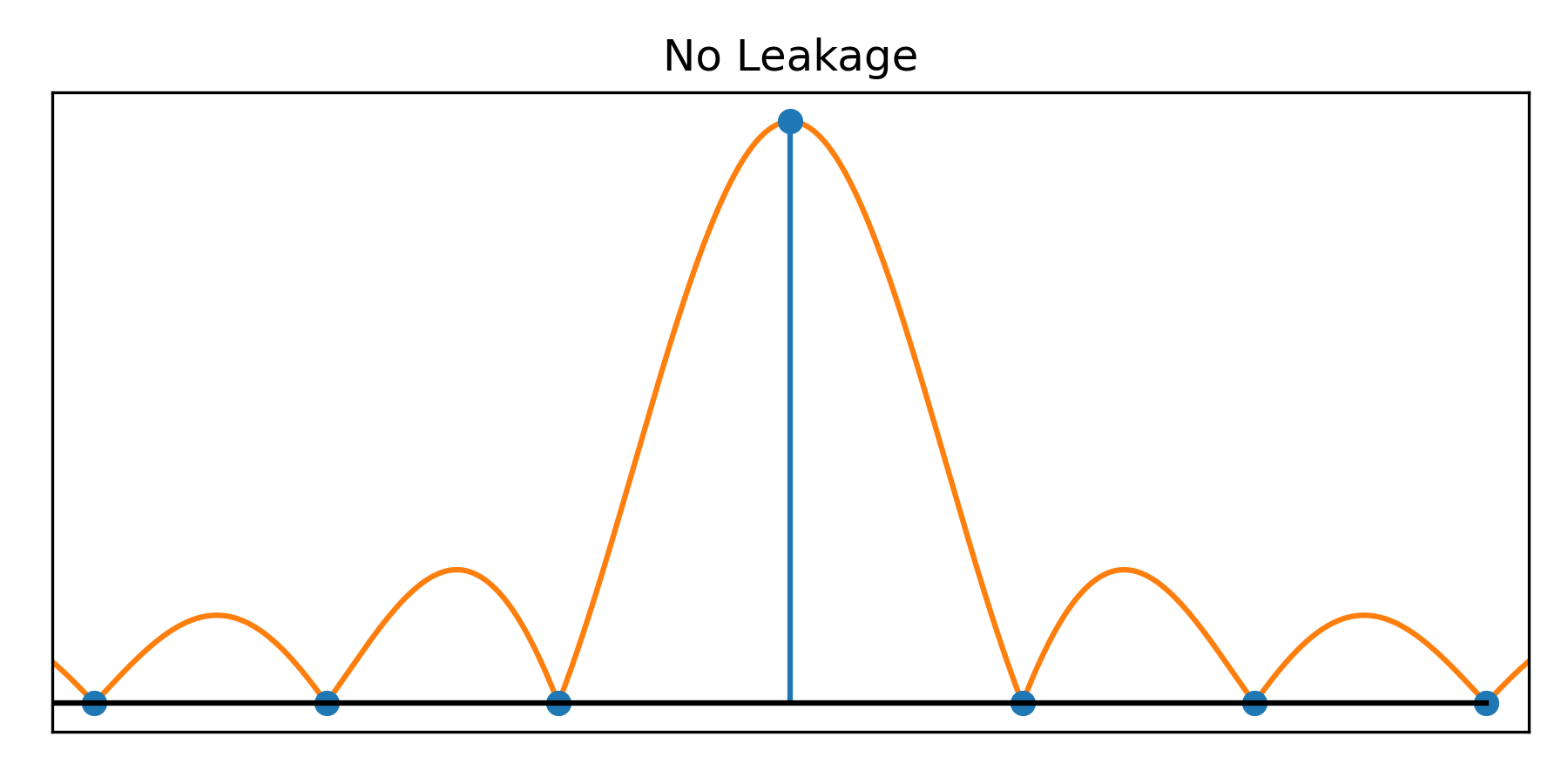}
    \caption{When the $\alpha = \beta_k$, there is no leakage}
    \label{fig:dirichlet_kernel_no_leakage}
  \end{minipage}
  \hspace{0.04\linewidth}
  \begin{minipage}[b]{0.45\linewidth}
    \centering
    \includegraphics[width=\linewidth]{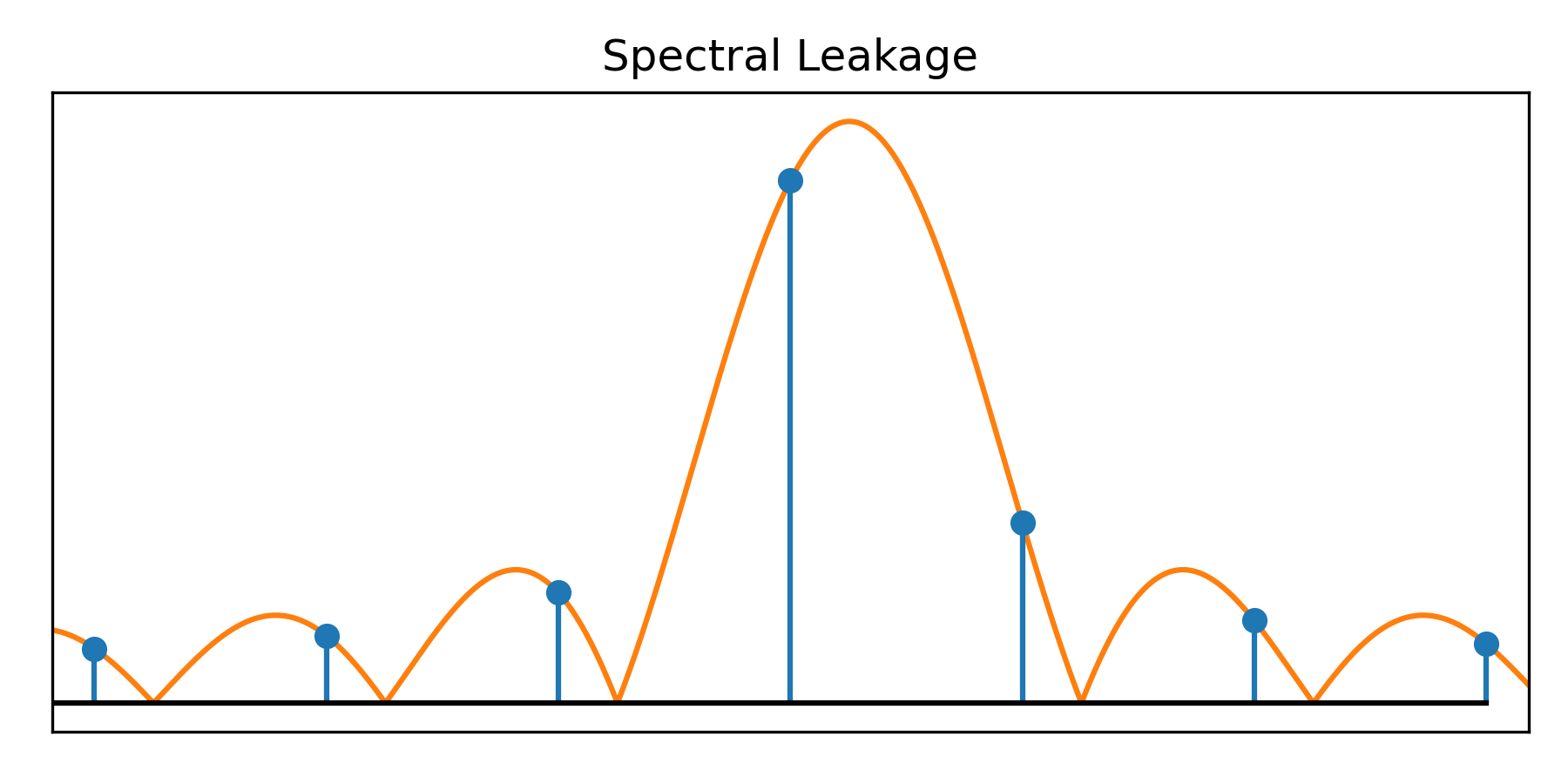}
    \caption{When the $\alpha \neq \beta_k$, there is leakage and in neighbouring bins}
    \label{fig:dirichlet_kernel_with_leakage}
  \end{minipage}
\end{figure}

This phenomenon—spectral leakage—introduces amplitude and phase spillage into adjacent bins as shown in Figures~\ref{fig:dirichlet_kernel_no_leakage} and ~\ref{fig:dirichlet_kernel_with_leakage}, reducing contrast and corrupting accurate frequency estimation. It is a key limitation in finite-length DFT analysis, especially relevant in radar signals where precise range estimation depends on spectral sharpness.

\subsection{Signal Synthesis}
\label{sec:forward_model}

In this section, we present the frequency-domain forward model that serves as the basis for our implicit neural representation framework for volumetric reconstruction. Instead of modeling the time-domain signal and applying a DFT afterward, we directly model the complex frequency-domain response at each DFT bin, leveraging the closed-form DFT derivation introduced in Section~\ref{DFT}. This formulation is fully differentiable and thus amenable to gradient-based optimization used in neural rendering pipelines. Moreover, it encapsulates physical wave propagation, including transmission and backscattering, within the radar's operating bandwidth.

Let $\mathbf{x} \in \mathbb{R}^3$ denote a 3D point in the scene, and let $\sigma(\mathbf{x}) \in \mathbb{R}$ represent the reflectivity or scattering strength at that point. Define:
- $\mathcal{X} \subset \mathbb{R}^3$: the bounded spatial domain of interest,
- $\mathbf{o}_T, \mathbf{o}_R \in \mathbb{R}^3$: positions of the transmitter and receiver,
- $R_T(\mathbf{x}) = \|\mathbf{o}_T - \mathbf{x}\|$: propagation distance from Tx to $\mathbf{x}$,
- $R_R(\mathbf{x}) = \|\mathbf{o}_R - \mathbf{x}\|$: propagation distance from $\mathbf{x}$ to Rx,
- $\tau(\mathbf{x}) = (R_T(\mathbf{x}) + R_R(\mathbf{x}))/c$: total round-trip delay.

Then, the complex response at DFT bin index $k$ is given by:
\[
\boxed{
Z_k = \int_{\mathcal{X}} \frac{\sigma(\mathbf{x})}{N R_T(\mathbf{x}) R_R(\mathbf{x})} \cdot e^{i \phi(\mathbf{x})} \cdot \frac{1 - e^{i 2\pi S \tau(\mathbf{x}) N}}{1 - e^{i (2\pi S \tau(\mathbf{x}) - \beta_k)}} \, d\mathbf{x}
}
\]
where:
- $\beta_k = \frac{2\pi k}{N}$ is the angular frequency of the $k$-th DFT bin,
- $\phi(\mathbf{x}) = 2\pi f_0 \tau(\mathbf{x})$ is the baseband phase shift due to initial frequency $f_0$,
- $S$ is the chirp slope ($S = B/T_c$),
- $N$ is the number of time-domain samples (or DFT points).Given ground truth measurements $\tilde{Z}_k$, the loss is defined as:

\[
\mathcal{L} = \sum_k \left\| \left| Z_k \right| - \left| \tilde{Z}_k \right| \right\|_2^2 + \lambda \sum_k \left( \left\| \Re(Z_k) - \Re(\tilde{Z}_k) \right\|_2^2 + \left\| \Im(Z_k) - \Im(\tilde{Z}_k) \right\|_2^2 \right)
\]

where $\lambda = 0.5$ controls the relative importance of magnitude vs. complex component supervision.

When $f_0 = 0$, the carrier-induced phase term vanishes, i.e., $\phi(\mathbf{x}) = 0$, simplifying the model without impacting range resolution (as established in prior sections). This forward model represents the coherent summation of returns from all scene points, modulated by the reflectivity $\sigma(\mathbf{x})$, and the frequency response derived from the DFT of a complex exponential tone with a delay $\tau(\mathbf{x})$.

This formulation can be extended to incorporate additional wave propagation phenomena, such as:
- Transmission attenuation models (e.g., based on material absorption),
- Scattering probability as a function of incident angle or local geometry (e.g., Lambertian, specular, or volumetric models),
- Multipath interference or occlusion. These extensions can be embedded as additional multiplicative factors within the integrand.





\section{Experimental Setup}
\label{sec:experiment_setup}

We evaluate our frequency-domain forward model using simulated FMCW radar data over a cylindrical synthetic aperture. The setup mimics a practical configuration where the object rests on a turntable and the radar sensor is mounted on a vertical actuator. Their combined motion produces a dense 3D sampling of the scene from multiple viewpoints, forming a cylindrical inverse synthetic aperture. Our radar simulation follows the TI AWR1843BOOST MIMO configuration, with a $3.585$~GHz bandwidth, $70.295 \times 10^{12}$~Hz/s chirp slope, and a sampling rate of $5$~MHz, yielding $256$ ADC samples per chirp. These time-domain beat signals are transformed via DFT into 256 frequency bins as discussed in Section~\ref{DFT}. The synthetic scene spans a $0.36$~m cube, with the radar positioned $0.23$~m from the object center. To obtain a well-structured dataset for learning, we first simulate $691{,}200$ multistatic MIMO radar measurements, then apply a multistatic-to-monostatic transformation to normalize Tx-Rx geometries and simulate a consistent forward model view. This sampling strategy closely mirrors real-world volumetric setups such as AirSAS~\cite{cowen2021airsas}, and provides a feasible path toward hardware realization compared to more idealized setups like spherical apertures. We train our models on RTXA5000 for 1500 epochs and batchsize of 1024 measurements.

\section{Evaluation}
\label{sec:evaluation}

We evaluate our frequency-domain forward model for volumetric reconstruction by comparing it with multiple baseline approaches that vary in terms of signal representation (time vs. frequency domain) and physical modeling fidelity. All methods are implemented in our volumetric neural reconstruction pipeline using the same cylindrical scanning geometry described in Section~\ref{sec:experiment_setup}.

\subsection{Baseline Methods}

(1) Time-domain forward model with Temporal Supervision (TF-TS):
This baseline implements a physically accurate time-domain forward model based on path delays. The loss is computed between the simulated and predicted signals in the time domain, i.e., before applying any frequency transform. \\
(2) Time-domain forward model with Spectral Supervision (TF-SS):
This method uses the same time-domain forward model as the baseline, but computes the loss after applying an FFT to convert the predicted signal into the frequency domain. This hybrid formulation partially accounts for spectral leakage and DFT artifacts.\\
(3) Range Quantization (RQ) model:
Inspired by traditional radar signal processing techniques, this method assigns each scatterer’s contribution to a DFT frequency bin based on its round-trip distance. The bin index is determined via quantized range thresholds, and scatterer contributions are summed into these bins to form a synthetic frequency-domain signal. While computationally efficient, this approximation ignores complex phasor interactions and introduces quantization artifacts. Loss is computed directly in the frequency domain.\\
(4) Coherent backprojection:
This traditional imaging method discretizes the 3D scene into voxels and estimates each voxel's intensity by coherently summing complex contributions from all Tx. Each contribution is phase-aligned based on the voxel's round-trip delay.

\subsection{\systemname{}}
Our method directly models the radar response in the frequency domain using a differentiable, closed-form expression derived from the DFT of a delayed tone (Section~\ref{sec:forward_model}). This enables more accurate modeling of spectral leakage, non-bin-centered frequencies, and phase information, while being fully compatible with gradient-based optimization. The forward model is used in conjunction with a neural volumetric representation and optimized end-to-end.



...

\begin{figure}[h]
  \centering

  \noindent\begin{tabularx}{\linewidth}{*{6}{>{\centering\arraybackslash}X}}
    {\small Range Quant.} & {\small TF-TS} & {\small TF-SS} & {\small Backprojection} & {\small SpINR (Ours)} & {\small Ground Truth} \\
  \end{tabularx}

  \vspace{2pt}

  \includegraphics[width=\linewidth]{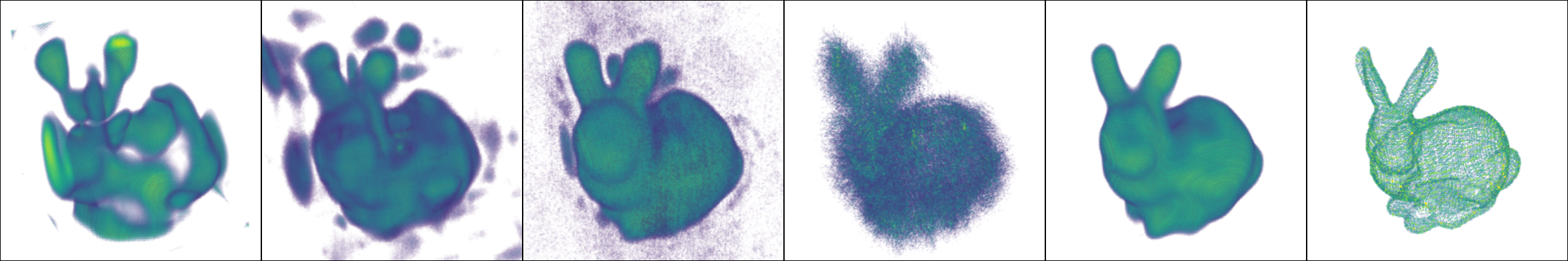}

  \includegraphics[width=\linewidth]{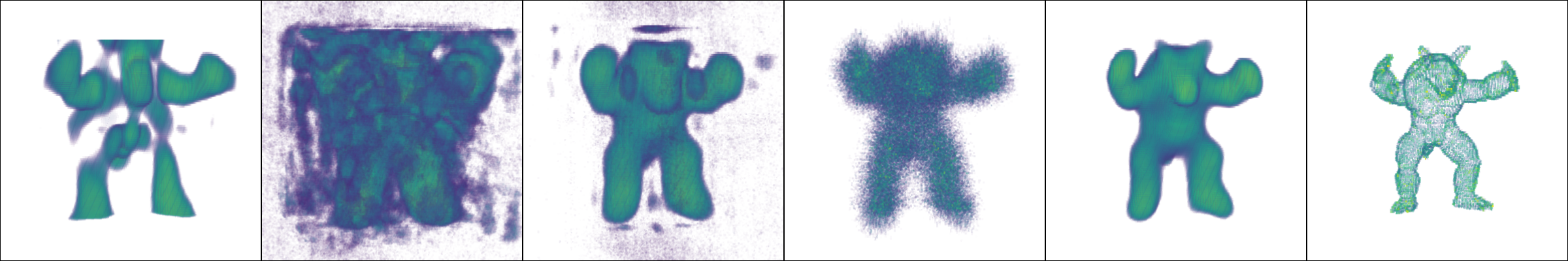}

  \includegraphics[width=\linewidth]{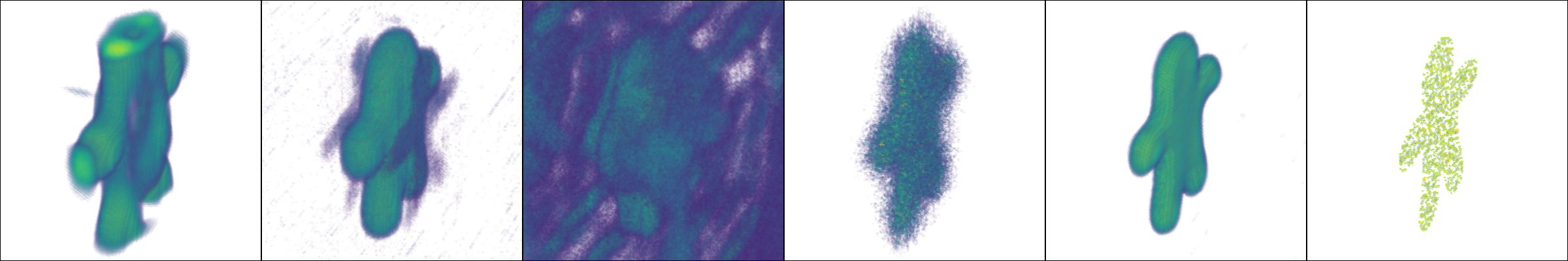}

  \caption{Comparison of volumetric reconstructions for Range Quantization, Time-domain forward model with Temporal Supervision, Time-domain forward model with Spectral Supervision, and our method. Our reconstructions more accurately match the ground truth geometry compared to other methods.}
  \label{fig:comparison}
\end{figure}

\subsection{Frequency domain Supervision vs. Time domain Supervision}
As explained in the previous section, frequency domain signal does range compression so there is a linear mapping, direct mapping between bin number and scatterer distance. Because the forward model is linear in that domain, the loss surface is simpler in terms of the true signal, making optimization more straightforward and physically meaningful than a purely image/time-domain loss.
 For instance, a slight shift in the scatterer location can cause time shift of the time domain waveform and lead to a large time-domain MSE, but its frequency spectrum remains nearly unchanged. As a result, the reconstruction from the frequency domain loss is much better than time domain loss as shown in Fig. \ref{fig:comparison}


\subsection{Frequency-Domain vs. Time-Domain Forward Model}
\label{sec:freq_vs_time}

\begin{figure}[h]
  \centering
  \begin{minipage}[b]{0.45\linewidth}
    \centering
    \includegraphics[width=\linewidth]{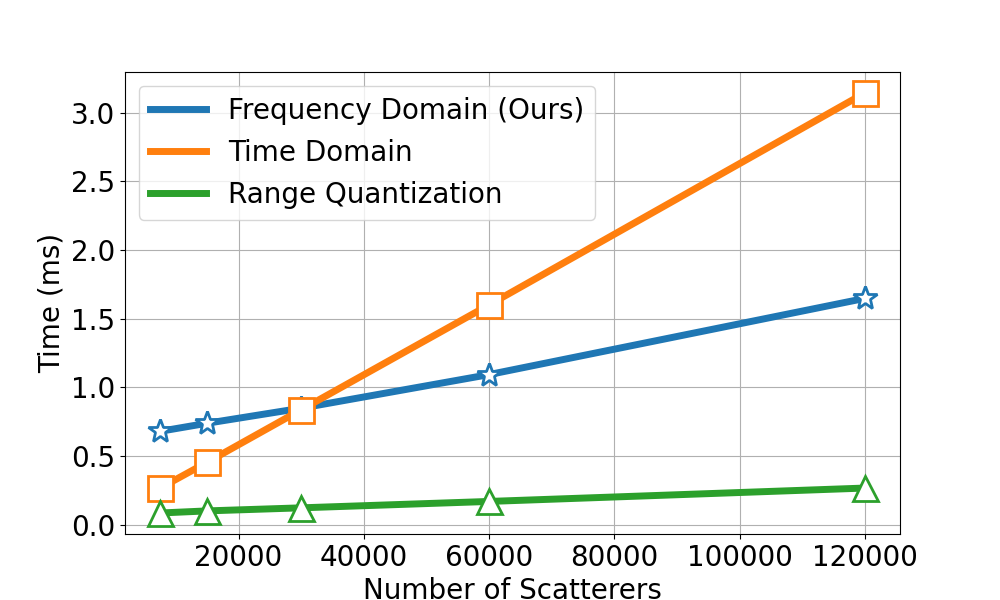}
    \caption{We measure the runtime of different forward models with different scene size governed by the number of scatterers. Our proposed method outperforms the time domain forward model while range quantization is the fastest.}
    \label{fig:forward_model_performancev2}
  \end{minipage}
  \hspace{0.04\linewidth}
  \begin{minipage}[b]{0.45\linewidth}
    \centering
    \includegraphics[width=\linewidth]{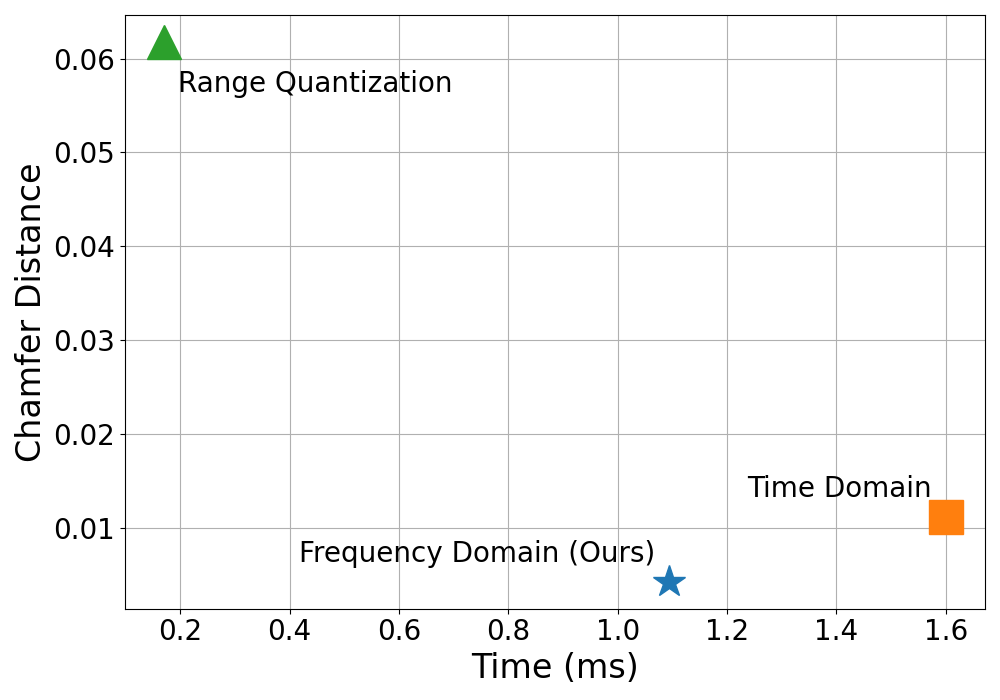}
    \caption{We compare the three forward models for their runtime and volume reconstruction accuracy. Even though the Range Quantization has the fastest runtime, it severe underperforms in volume reconstruction as signified by the chamfer distance (lower better).}
    \label{fig:Fdfwd_vs_Tdfwd2}
  \end{minipage}
\end{figure}

Since the maximum range of our scene is known and bounded, we can precompute the number of frequency (range) bins required to fully capture the signal energy. In our case, this corresponds to only 16 frequency bins that fall within the observable round-trip delays from the scene. This provides a key computational advantage for our proposed frequency-domain forward model: we compute only the 16 required frequency components directly using our closed-form expression (as derived in Section~\ref{sec:forward_model}), bypassing unnecessary intermediate steps. In contrast, the time-domain forward model must simulate the full time-domain response over 256 samples, then apply an FFT to obtain the 256-point frequency-domain signal, and finally truncate it to retain just the 16 relevant bins for loss computation. This leads to significant additional overhead. To quantify this efficiency gap, we compare the average runtime per forward pass of both models on an NVIDIA RTX A5000 GPU. As shown in Fig~\ref{fig:forward_model_performancev2}, our frequency-domain model achieves lower latency by avoiding unnecessary computation. This runtime improvement, combined with the differentiability and spectral fidelity of our formulation, makes the frequency-domain model well-suited for learning-based volumetric reconstruction.



\subsection{Why Frequency-Domain Forward Modeling Improves Learning Stability}
\label{sec:gradient_stability}

Although both our proposed frequency-domain forward model and the time-domain baseline ultimately compute loss in the frequency domain, we observe a significant performance gap in reconstruction accuracy and convergence speed. To investigate this, we analyze the gradient dynamics of the two approaches and show that our method yields smoother, more stable gradients throughout the network—enabling faster and more effective optimization.

\paragraph{Motivation from Prior Work.}
The importance of gradient flow in neural network design is well established. Deep architectures often suffer from vanishing or exploding gradients, motivating skip connections~\cite{10.5555/3305381.3305417}, normalized initialization~\cite{zhang2019fixup}, and sinusoidal activations for signal representation~\cite{10.5555/3495724.3496350}. These works demonstrate that preserving coherent and stable gradients across layers is crucial for trainability, especially in models with many layers or complex signal mappings. Similar reasoning has been used to justify architectural choices in implicit neural representations~\cite{10.5555/3495724.3496350}, and sparse pruning strategies~\cite{2002.07376}, Following this line, we examine how the formulation of the forward model affects gradient propagation in neural volumetric reconstruction.

\begin{figure}[h]
    \centering
    \subfigure[Mean]{
        \includegraphics[width=0.45\linewidth]{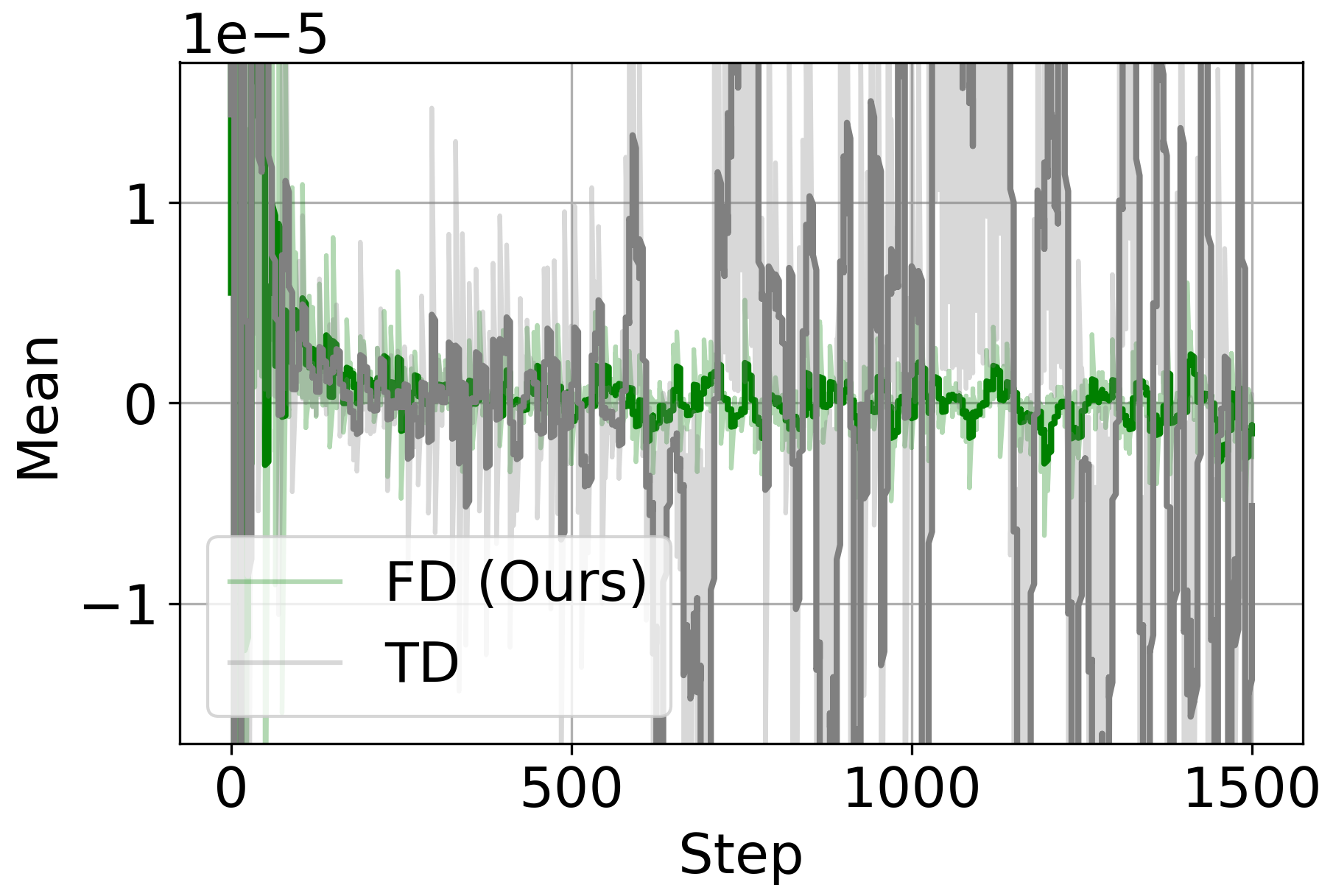}
    }%
    \subfigure[Standard Deviation]{
        \includegraphics[width=0.45\linewidth]{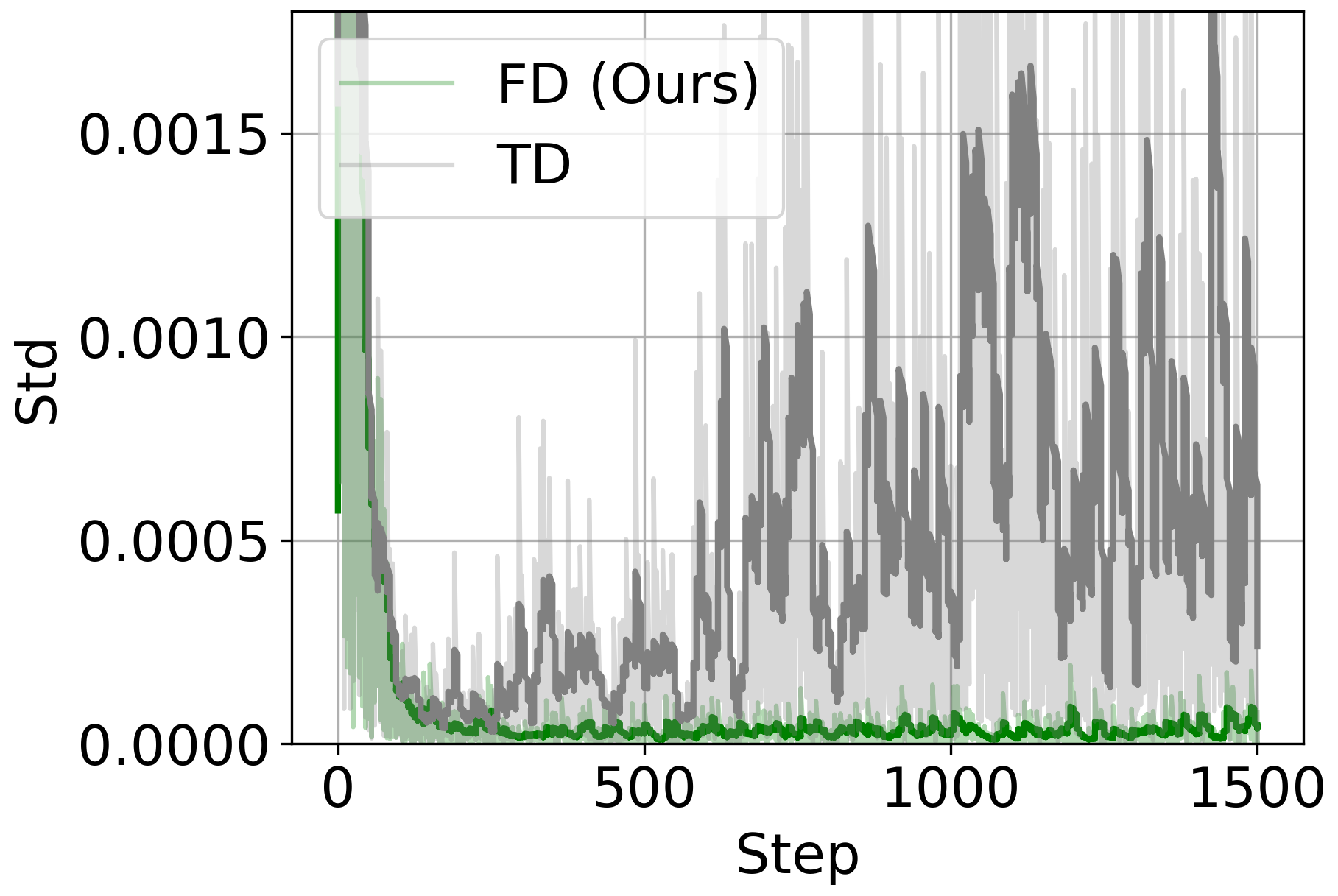}
    }\\
    \caption{We log the (a) mean and (b) standard deviation of the gradients for the first layer wrt the number of training steps. The mean and standard deviation are more stable for our proposed method. For the time domain forward model the the gradient tends to explode.  We observe a similar trend for all the subsequent layers.}
    \label{fig:gradient_experiment}
\end{figure}

We visualize the computation graph of both forward models (Figure~\ref{fig:gradient_experiment}) and observe that the time-domain model includes additional operations (e.g., time-domain simulation, full FFT) that introduce longer, more fragmented backpropagation paths. In contrast, our frequency-domain forward model analytically predicts only the relevant frequency bin values, resulting in a shallower and more direct gradient path to the implicit neural representation. To quantify this, we log the mean and standard deviation of the weight gradients at the first layer of the INR across training. Our method (green) exhibits significantly more stable gradient statistics than the time-domain model (gray). The variance remains controlled throughout training, avoiding sudden spikes or vanishing behavior. 

These findings mirror prior observations from studies on residual networks~\cite{10.5555/3305381.3305417}, gradient confusion~\cite{10.5555/3524938.3525723}, and signal representations~\cite{10.5555/3495724.3496350}, where architectural decisions that preserve or align gradients lead to better optimization and generalization. Our work extends this perspective to signal modeling for FMCW radar, showing that a forward model grounded in spectral domain physics not only improves interpretability but also facilitates learning through more stable and effective gradient flow.

\subsection{Our Frequency-Domain Model vs. Classical Coherent Backprojection}
\label{sec:vs_backprojection}

To further validate the effectiveness of our proposed frequency-domain forward model, we compare it against a classical reconstruction baseline: coherent backprojection (CBP). This method discretizes the scene into voxels and accumulates backprojected energy from all Tx paths by aligning phases based on path delays. While physically grounded, CBP has significant limitations that our learning-based approach addresses. \\
(1) Inherent Discretization and Aliasing:  
CBP relies on uniform voxel grids and discrete binning of round-trip delays, which often leads to aliasing artifacts and blurring, especially when the sampling aperture is non-uniform or sparse. Our method, by contrast, uses a continuous volumetric representation (via neural fields) and avoids explicit voxelization, enabling sub-voxel accuracy in both representation and rendering. \\
(2) Lack of Differentiability and Learning:  
CBP is a purely geometric method—it does not learn from data or optimize a forward model. As a result, it is highly sensitive to aperture coverage, noise, and partial views. Our method incorporates a fully differentiable, physics-informed forward model, enabling gradient-based optimization that refines the scene representation to best match all measurements.


As shown in our quantitative results (Tables~\ref{tab:evaluation},~\ref{tab:chamfer_distance}), \systemname{} achieves significantly better reconstruction compared to CBP.




\section{Quantitative Evaluation}
\label{sec:quant_evaluation}

We evaluate the quality of 3D volumetric reconstruction across five methods using standard metrics widely adopted in computer vision and inverse rendering. These include Intersection-over-Union (IoU), Chamfer Distance (CD), Hausdorff Distance (HD), Peak Signal-to-Noise Ratio (PSNR), Structural Similarity Index (SSIM), and Learned Perceptual Image Patch Similarity (LPIPS). Together, they capture both geometric fidelity and perceptual image quality by comparing voxel grids, rendered views, and point cloud representations of reconstructed shapes.

We compare our method, \systemname{}, against: (i) a time-domain forward model with time-domain loss (TF-TS), (ii) a time-domain forward model supervised in the frequency domain (TF-SS), (iii) a reconstruction-by-querying baseline (RQ), and (iv) classical backprojection. All methods are trained on the same synthetic cylindrical radar dataset described in Section~\ref{sec:experiment_setup}.

\textbf{Quantitative Results.} Table~\ref{tab:evaluation} summarizes average performance across all test scenes. Our model achieves the best performance across all six metrics, showing clear improvements in both geometric reconstruction (IoU, CD, HD) and view-based similarity (PSNR, SSIM, LPIPS). Table~\ref{tab:chamfer_distance} presents Chamfer Distance per object, where \systemname{} outperforms all baselines with consistent gains across diverse shapes.

\begin{table}[h]
\centering
\begin{tabular}{|c|c|c|c|c|c|c|}
\hline
Method & IoU ($\uparrow$) & Chamfer ($\downarrow$) & Hausdorff ($\downarrow$) & PSNR ($\uparrow$) & SSIM ($\uparrow$) & LPIPS ($\downarrow$) \\ 
\hline
\systemname{} & \textbf{0.0908} & \textbf{0.0055} & \textbf{0.0713} & \textbf{17.06} & \textbf{0.801} & \textbf{0.248} \\
TF-TS          & 0.0483          & 0.0219          & 0.1470          & 9.27           & 0.469          & 0.683 \\
TF-SS          & 0.0484          & 0.0298          & 0.1862          & 7.35           & 0.379          & 0.776 \\
RQ            & 0.0135          & 0.0728          & 0.1856          & 6.27           & 0.390          & 0.806 \\
Backproj      & 0.0598          & 0.0099          & 0.0461          & 11.28          & 0.691          & 0.426 \\
\hline
\end{tabular}
\caption{Average reconstruction performance across all scenes.}
\label{tab:evaluation}
\end{table}

\vspace{-2mm}
\begin{table}[h]
\centering
\begin{tabular}{|c|c|c|c|c|c|c|c|}
\hline
Method & bunny & spot & lucy & armadillo & dragon & woody & teapot \\ 
\hline
\systemname{} & \textbf{0.0050} & \textbf{0.0066} & \textbf{0.0044} & \textbf{0.0042} & \textbf{0.0042} & \textbf{0.0061} & \textbf{0.0080} \\
TF-TS          & 0.0111          & 0.0594          & 0.0055          & 0.0196          & 0.0056          & 0.0110          & 0.0408 \\
TF-SS          & 0.0085          & 0.0471          & 0.0077          & 0.0065          & 0.0059          & 0.0938          & 0.0390 \\
RQ            & 0.0593          & 0.0859          & 0.0690          & 0.0618          & 0.0625          & 0.0671          & 0.1041 \\
Backproj      & 0.0095          & 0.0160          & 0.0074          & 0.0079          & 0.0072          & 0.0090          & 0.0124 \\
\hline
\end{tabular}
\caption{Chamfer Distance for each object (lower is better).}
\label{tab:chamfer_distance}
\end{table}

\section{Conclusion}
In this paper, we introduced \systemnamesansours{}, a novel framework for volumetric reconstruction using Frequency-Modulated Continuous-Wave (FMCW) radar data. By integrating a fully differentiable forward model operating in the frequency domain with implicit neural representations (INRs), SpINR effectively leverages the inherent linear relationship between beat frequency and scatterer distance. This approach not only facilitates more efficient and accurate learning of scene geometry but also enhances computational efficiency by focusing on relevant frequency bins.

Our extensive experiments demonstrate that \systemnamesansours{} significantly outperforms classical backprojection methods and existing learning-based approaches, achieving higher resolution and more accurate reconstructions of complex scenes. This advancement represents the first application of neural volumetric reconstruction in the radar domain, offering a promising direction for future research in radar-based imaging and perception systems.

Future work could explore extending SpINR to dynamic scenes, incorporating motion estimation to handle moving objects. Additionally, integrating SpINR with other sensor modalities, such as LiDAR or optical cameras, could further enhance reconstruction accuracy and robustness. 

\section{Acknowledgement}
This work was partially supported by NSF CAREER Award 2238433. We also thank the various companies that sponsor
the iCoSMoS laboratory at UMD.



\clearpage

\end{document}